# Application of 2-D Convolutional Neural Networks for Damage Detection in Steel Frame Structures

Shahin Ghazvineh[1], Gholamreza Nouri[1], Seyed Hossein Hosseini Lavassani[1], Vahidreza Gharehbaghi[2], Andy Nguyen[2]

*[1] Faculty of Civil Engineering, Kharazmi University, Tehran, Iran*
*[2] University of Southern Queensland, Springfield Campus, Queensland, Australia*

## Abstract

Extracting damage-sensitive features that are insensitive to noise from time-series data was a significant challenge in traditional machine learning-based structural damage detection (SDD) methods. Moreover, this procedure slowed down the techniques and made the methods' accuracy dependent on preprocessing. By growing deep learning (DL), several studies were conducted to solve this limit using deep architectures in the field of structural health monitoring (SHM). However, these methods mostly require substantial measurements for the training procedure and are expensive to process, making them unsuitable for real-time approaches. Thus, we present an application of 2-D convolutional neural networks (2-D CNNs) designed to perform both feature extraction and classification stages as a single organism to solve the highlighted problems. The method uses a network of lighted CNNs instead of deep and takes raw acceleration signals as input. Using lighted CNNs, in which every one of them is optimized for a specific element, increases the accuracy and makes the network faster to perform. Also, a new framework is proposed for decreasing the data required in the training phase. We verified our method on Qatar University Grandstand Simulator (QUGS) benchmark data provided by Structural Dynamics Team. The results showed improved accuracy over other methods, and running time was adequate for real-time applications.

# 1 Introduction

Environmental, operational, and other human impact elements make structural damage to civil structures unavoidable. In order to minimize structural degradation and collapse and improve serviceability and preserve human lives, it is necessary to inspect civil structures on a regular schedule due to the progressive nature of the damage. As a result, structural damage detection (SDD) is becoming a popular research topic among civil engineering academics and practitioners. [1–3].

There are two types of structural damage detection techniques: local and global [4]. Due to their ability to identify, measure, and locate structural damage by processing accelerometer signals, research studies on vibration-based damage detection methods [5–8] have increased in quantity as a sub-area of global detection techniques. Model-based (parametric) and signal-based (nonparametric) vibration-based strategies exist [4, 9]. Modal parameters (i.e. natural frequency and mode shape), damping, and stiffness indices vary when damage occurs using parametric SDD approaches. Nonparametric approaches, on the other hand, can detect damage from observed signals even if the structural model is unknown [10].

A combination of autonomous data analysis and the growing data size has made machine learning (ML) approaches extremely useful in the SDD field. Researchers have used a variety of machine learning techniques to keep tabs on the health of a structure as a consequence. Preprocessing measured data, extracting handcrafted features from it, and categorizing features with a classifier to reveal whether or not the structure is healthy or damaged are all steps in an ML-based SDD technique. Since traditional ML approaches require manual damage-sensitive features, this poses a significant difficulty because it is challenging to implement in practice. Most of the time, the performance of machine learning methods is determined by the quality of these handcrafted characteristics, which must be obtained through a lengthy process of trial and error. The development of an automated feature extraction method is required since feature extraction is critical in detecting damage to civil constructions [11].

A subclass of ML called deep learning (DL) emerged with the rise in the computational capacity of computers, becoming a viable solution in several fields as computer vision (12,13), speech recognition (14), and signal processing (15,16,17). The problem has been solved since the DL models' auto-extracting capability eliminates the need for manual features [20]. The convolutional neural network (CNN) is a common deep learning (DL) model that can extract features from raw input and learn them with the classifier in a single learning framework. As a result, researchers in the field of SDD have presented a large variety of CNN-based techniques.

According to Modarres et al. [21], the proposed methodology was tested against four existing methods for identifying noise and damage in images, and the findings showed that the new method outperformed the others by a wide margin. Both instances saw better accuracy and precision with 2-D CNN, according to the results. A similar 2-D CNN design for identifying wind turbine blade surface damage was proposed by Shihavuddin et al. Even though 2-D CNNs were designed for image-based inputs in nature, vibration-based SDD methods are being used. These methods use 2-D CNN by changing the input data format from signal to image [10], by using strain data in a 2-D arrangement [23], or by concatenating the acceleration or dynamic displacement responses measured by a network of sensors into a 2-D matrix [24,25] as input to benefit from the 2-1dD form of CNNs. Signal processing, on the other hand, frequently makes use of 1D-CNN architectures. Research has thus far been done on the use of 1D-CNNs for SDD problems employing time-domain acceleration signals for different types of challenges. The first mode shape of the structure [32] and frequency-domain responses [31] are two examples of alternative input data types that have been used in research. Adding extra convolution layers to their models creates a complex architecture that requires a lot of input and is computationally expensive to process. Using only two states of the structure (completely damaged and undamaged data) as input for the training stage, Avci et al. [33] developed a decentralized 1-D CNN-based technique to estimate structural damage. Avci et al. [34] presented a decentralized SDD technique for wireless sensor networks based on 1-D CNN. To do this, a separate network was created for each sensor to keep track of whether or not the corresponding joint in a steel frame was damaged. While the method was computationally inefficient, the results were impressive in terms of running time. However, in terms of accuracy, there is still room for improvement.

As a result, the authors suggest a novel vibration-based SDD method based on two-dimensional CNN for detecting and localizing structural degradation in civil constructions. The method is deemed decentralized because it does not attempt to prepare or engineer features into the raw acceleration data. We sought to develop a highly accurate method that requires less time for new sample evaluation and can be trained with fewer data points than previous methods. By decentralizing, we may improve accuracy by optimizing each CNN for each element in a structure while maintaining a compact and lightweight architecture. As a result, computational complexity is lowered, and each CNN may be completed faster. To assess the suggested method's viability, the authors trained a basic two-dimensional CNN for each joint in the Qatar University Grandstand Simulator (QUGS) data set [35–37], utilizing acceleration data from three nearest accelerometers to each joint as input. Three different degrees of white Gaussian noise (WGN) are also introduced to the data sets in the second round of the evaluation to investigate the method's capacity to learn in noisy situations.

The following is the paper's outline. To begin, the proposed technique is described, along with a quick overview of CNN, the training, and the test approach. The following part discusses the benchmark problem and explains how the proposed solution is applied. Finally, based on the results, conclusions and feature works are discussed.

# 2 Methodology

This section examines the use of 2-D CNNs for SDD techniques in order to develop a robust, accurate, and rapid algorithm that does not require a massive quantity of training data. The authors offer a decentralized strategy for accomplishing this goal. The primary reason for utilizing the 2-D variant of CNN in this work is that, unlike the 1-D kind, 2-D CNN kernels move in two directions. This capacity, together with the addition of more learnable characteristics, enhances learning capability. Utilizing 2-D CNNs also enables us to reduce the amount of data required for the training phase by altering the input-form, as discussed in the training section. The next subsections describe the architecture and a brief description of each layer of the proposed CNNs, followed by explaining the training and testing methodologies.

## 2.1 CNN Architecture

In the realm of SDD, it is typical to train a single CNN for multi-class classification. Typically, these approaches should utilize a deep architecture of the CNN by adding more convolutional blocks to enable detection of defects throughout the structure, which adds computing complexity and raises the requirement for training data. The authors propose employing a network of CNNs with simple configurations rather than a deep CNN to overcome the aforementioned issues. This approach allocates a unique CNN to each structural element in order to analyze its condition. Figure 1 illustrates the recommended methodology.

Each CNN's primary architecture is straightforward, consisting of two convolutional blocks followed by a dropout function before reaching the fully connected (FC) layer. Each convolutional block comprises the following layers: convolutional, batch normalization, activation, and pooling. Prior to the first convolution block, batch normalization is applied to the input data to reduce non-uniform distribution. The following section provides an overview of each CNN layer.

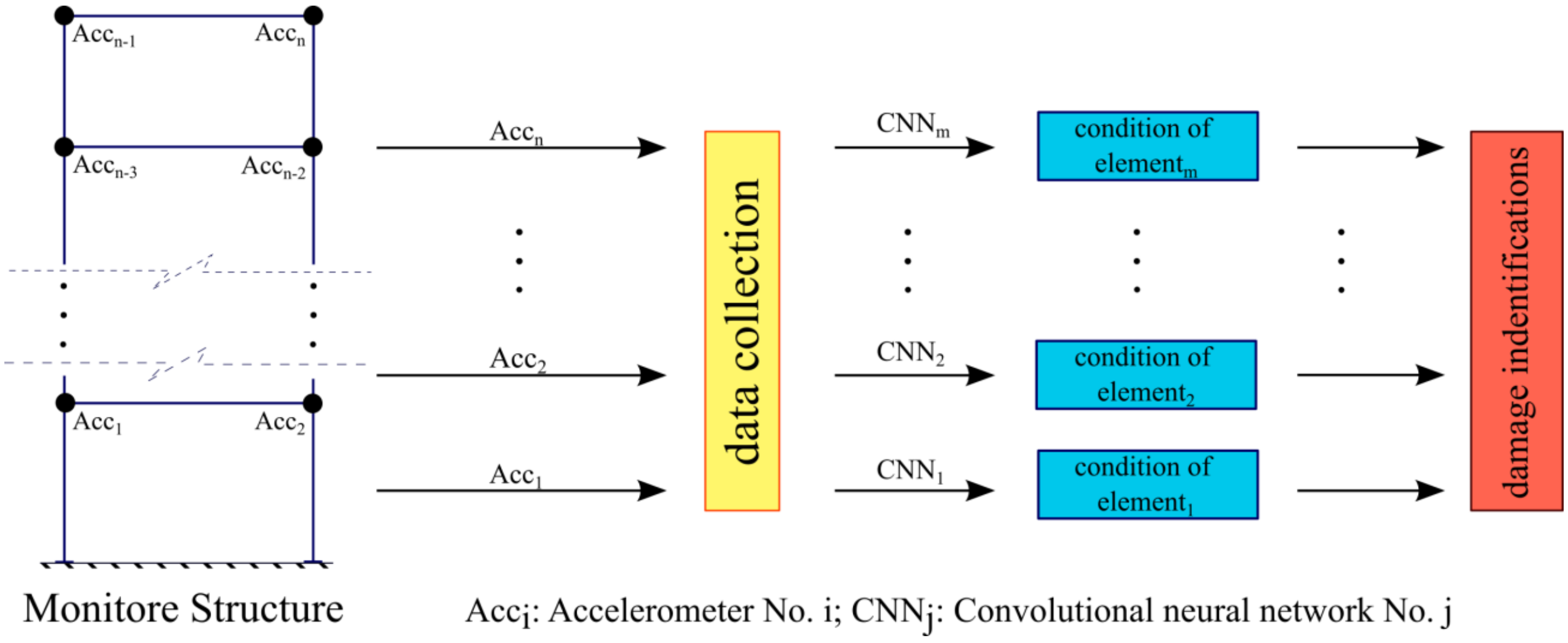

*Figure 1; The evaluation system of the proposed methodology*

### 2.1.1 Convolutional Layer

As implied by the name, the convolutional layer is what distinguishes CNNs from more standard machine learning approaches. Convolution is a mathematical procedure used in this layer to extract features from the layer's input. A convolution layer is composed of many kernels with learnable weights that convolve the input or previous layer's output one by one and produce a new 3D matrix called the feature map that is fed to the next layer.

### 2.1.2 Batch Normalization

Because the training process is performed batch by batch, non-uniformly distribution of every batch can extremely slow the convergence. To address this problem known as internal covariate shift, Sergey Ioffe and Christian Szegedy [38] proposed a method called batch normalization (BN). Batch normalization is an effective method for parametrizing almost all kinds of neural networks. It works by calculating the mean and variance of every input batch, then shifting and scaling them to 0 and 1, respectively. Since BN is performed on a mini-batch, it is computationally inexpensive, and finally, it helps to use bigger learning rates.

### 2.1.3 Activation Function ReLU

One of the most important components of a neural network is the activation functions. With their help, the network's input may learn and accomplish increasingly complicated tasks since they add non-linearity.

Most people are familiar with and comfortable with the rectified linear unit (ReLU) [39,40]. It's as simple as this: max (0,x). There are two advantages to using ReLU over other algorithms: first, it can deal with problems like the disappearing gradient that plagued sigmoid [41], and second, it uses less computing power, making it faster to train.

### 2.1.4 Max Pooling

A pooling layer follows the activation function as the final convolutional block layer in the proposed CNNs. It seeks to reduce the feature map's size by retaining only the most essential data in order to make it easier to feed-forward to the next layer. Max-pooling and average-pooling are the two most popular methods of pooling. Average-pooling slides a window through a feature map to determine the mean, while max-pooling grabs and keeps the highest value and discards the rest. We opted for the more efficient way of max-pooling for this project.

### 2.1.5 Dropout

For the first time, Nitish Srivastava et al. [42] introduced this strategy as a straightforward solution to the overfitting problem. In a nutshell, dropout ignores some units in the neural network during training with a probability of p, forcing the network to learn the features rather than memorize the training input data, which contains noise. Additionally, it reduces training time by lightening the network. Dropout is employed prior to the classifier layer in order to avoid CNNs from becoming overfit.

### 2.1.6 FC Layer

Before using a classifier, the features extracted using convolutional blocks must be flattened. FC layers use classification vectors, the architecture's last layers, to classify feature maps. It's worth noting that every neuron in an FC layer is linked to every activation in the one before it. Equation (1) demonstrates that each input value u is multiplied by a weight w and added to a bias b before being summed. Weight w and bias b are tunable parameters in this method. The FC layer generates an N-dimensional vector, where N is the number of classes in the problem, which in this case is 2.

$$y = \sum u \times w + b \tag{1}$$

### 2.1.7 Loss Function and Optimizer

For the calculation of the network's prediction error, Cross-Entropy [43] is employed as a loss function. If actual labels deviate from predictions, the Cross-Entropy loss function identifies this variation. Instead of using standard stochastic gradient descent (SGD), Adam optimizer method [44] is used to update network weights during training while decreasing the CNN's loss function value.

## 2.2 Training and Testing Procedure

As previously stated, this technique assesses the structure's condition by keeping track of each individual component using a unique 2-D CNN. Decentralizing, in contrast to most previous centralized investigations, aids in detecting various structural problems occurring at the same time. Raw acceleration signals under damaged and undamaged settings are used in the training of the CNNs. Three accelerometer signal data are used as input in the suggested approach. Reduce the amount of training data needed by using two extra accelerometer signals. With M elements and N accelerometers, for example, training a CNN to keep track of all the elements is necessary. Each accelerometer in the network records the acceleration reactions of the structure. In order to train all the CNNs, NZ features must be collected from the accelerometers, while adding signals from two additional existing accelerometers reduces it to roughly NZ/3 features. If Z features are required to train each CNN, only one accelerometer signal is required for each element. The authors proposed using supplemental signals from the element's closest accelerometers, which have more distinct characteristics. For element *i*, undamaged and damaged data are as follows:

$$U_i = [U_{i1} \quad U_{i2} \quad U_{i3}] \tag{2}$$

$$D_i = [D_{i1} \quad D_{i2} \quad D_{i3}] \tag{3}$$

The measurement signals $U_i$ and $D_i$ refer to the ith element's undamaged and damaged states, respectively. The acceleration responses recorded by the ith element's first, second, and third closest accelerometers are denoted by the terms $U_{i1}$, $U_{i2}$, and $U_{i3}$. The damaged state follows the same trend.

The undamaged data set for the ith element comprised signal responses in the ith element's healthy state as well, whereas one of the other elements was damaged.

For this reason, after collecting responses from both damaged and undamaged ith elements, the next step is to partition them into several fixed-length frames with a length of s f. Frames that have been damaged are designated with a 1, and those that have not been labeled with a 0.

$$U_i = \begin{bmatrix} U_{i1,1} & U_{i2,1} & U_{i3,1} \\ U_{i1,2} & U_{i2,2} & U_{i3,2} \\ & \vdots & \\ U_{i1,N_u} & U_{i2,N_u} & U_{i3,N_u} \end{bmatrix} \tag{4}$$

$$D_i = \begin{bmatrix} D_{i1,1} & D_{i2,1} & D_{i3,1} \\ D_{i1,2} & D_{i2,2} & D_{i3,2} \\ & \vdots & \\ D_{i1,N_d} & D_{i2,N_d} & D_{i3,N_d} \end{bmatrix} \tag{5}$$

Where $N_d$ and $N_u$ are the number of damaged and undamaged frames, respectively. In Equations 6 and 7, the *j*th frame of signals in the undamaged and damaged condition of element *i* is depicted. Each frame has a dimension of $s_f \times 3$.

$$UF_{i,j} = [U_{i1,j} \quad U_{i2,j} \quad U_{i3,j}] \tag{6}$$

$$DF_{i,j} = [D_{i1,j} \quad D_{i2,j} \quad D_{i3,j}] \tag{7}$$

Assuming that each accelerometer's signal is composed of $s_u$ and $s_d$ samples representing the undamaged and damaged states of an element, respectively, the total number of frames in the undamaged and damaged states is as follows:

$$N_u = \frac{s_u}{s_f} \tag{8}$$

$$N_d = \frac{s_d}{s_f} \tag{9}$$

Due to the fact that damage to other components is considered the healthy condition for a single element, the values of N u and N d are wildly dissimilar, creating an unbalanced data set for training. To overcome this issue, an equal number of rows from each scenario's undamaged condition is picked, separated into equal-length frames, and shuffled; the first N d frames are then used as undamaged frames for training.

Following the preparation of the relevant data for each CNN as described above, they should be trained independently using the backpropagation technique over numerous epochs. During the training phase, the best performance of each CNN will be saved based on validation data.

The final step of this method is to test each CNN on previously unseen data. The testing technique is as follows:

1. Divide input signals into frames of fixed-length $s_f$.
2. For each element, classify undamaged frames as 0 and damaged frames as 1 by the related CNN.
3. Calculate the percentage of the damage possibility ($DP_i$) for the *i*th element by averaging the outputs as follow:

$$DP_i = \frac{\sum_{j=1}^{n_i} L_{i,j}}{n_i} \times 100 \quad (10)$$

Where $L_{i,j}$ is the label output of the $j$th frame processed by $CNN_i$ and $n_i$ is the total number of the input frames classified by $CNN_i$. $DP_i$ is expected to be close to 0 for undamaged elements and 100 for damaged elements.

# 3 Experimental Validation On a Benchmark Problem

The purpose of this study is to assess the suggested approach against a novel SHM benchmark problem. The following parts describe the benchmark and then analyze the experimental validation of the current work. The final portion discusses the CNNs' speed.

## 3.1 QUGS Benchmark Problem

The Qatar University Grandstand Simulator, depicted in Figure 2, is a large-scale laboratory construction (QUGS). The benchmark is a steel structure supported on four columns by eight girders and 25 filler beams. This structure contains thirty accelerometers in thirty joints. Different damage scenarios are simulated by loosening the bolts at the beam-to-girder joints using a white noise shaker with a 1024 Hz sampling rate. Reference [37] contains further information about the benchmark structure. Structural Dynamics Team published the data as a benchmark on a website in 2018 [45].

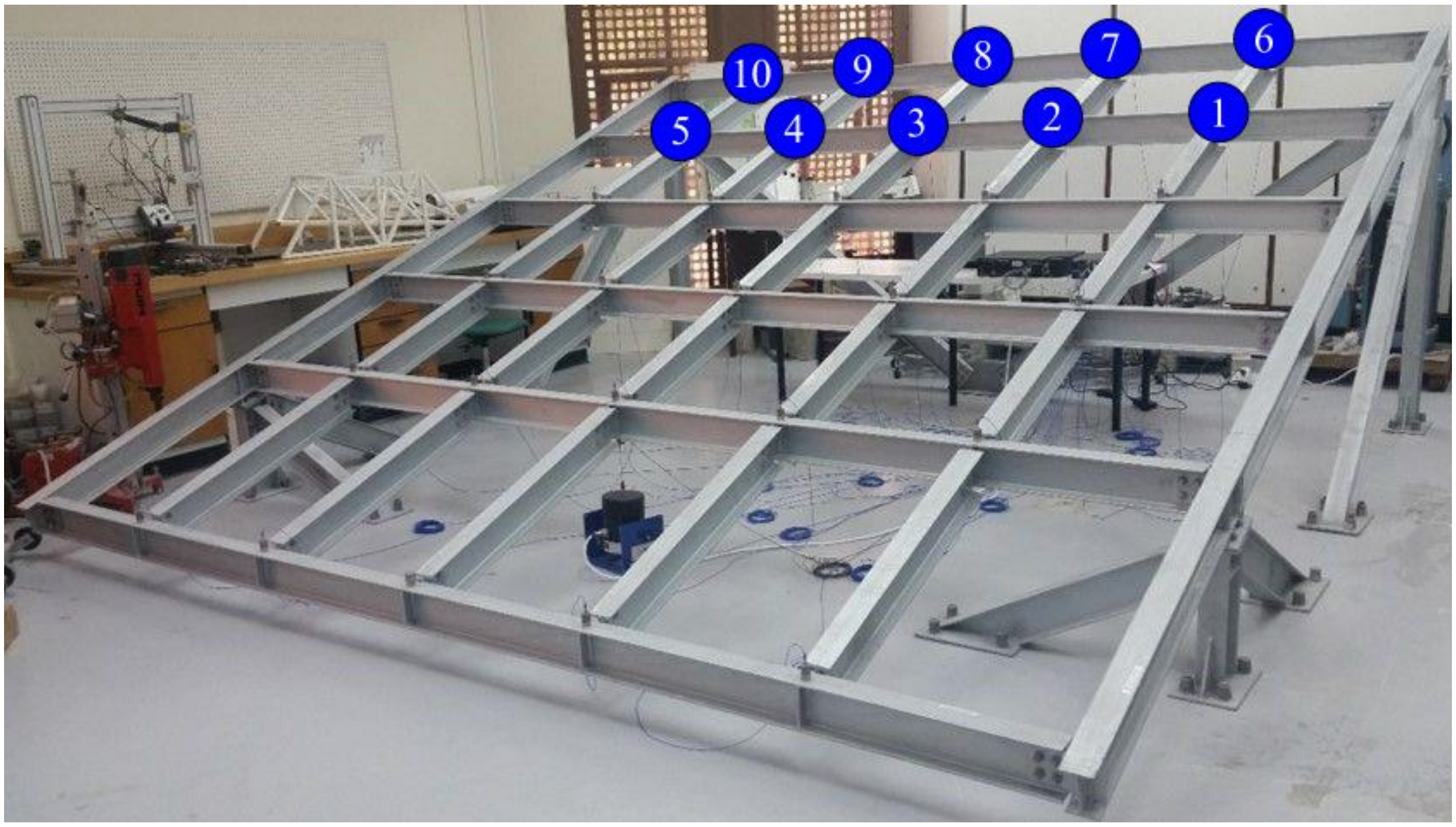

*Figure 2; Qatar university grandstand simulator structural steel frame (photo courtesy [33])*

## 3.2 Validation Results

For the initial attempt, just the top ten joints of the benchmark are tracked to determine the suggested methodology's efficacy. The data set for these 10 joints covers thirteen distinct situations, including ten single-joint injury instances, two double-joint damage cases, and one entire case. Each signal is composed of 262144 samples taken over 256 seconds. Half of the samples in data set A are split into fixed-length frames of length 256 and used to train ten 2-D CNNs using the technique outlined in Section 2.2, while the remaining 12.5 percent are used to test them. Additionally, 50% of data set B is used to validate the approach.

The selection of accelerometers is summarized in Table 1. Each CNN's hyper-parameters are chosen using a trial-and-error approach. CNNs were trained using the PyTorch framework in the Python 3.8 language environment, an Intel Core i7-4720HQ CPU and an NVIDIA GTX 950m graphics card to expedite the training process.

*Table 1; selection of the accelerometers for each joint*

| **Joint No.** | | | | | | | | | |
|---|---|---|---|---|---|---|---|---|---|
| 1 | 2 | 3 | 4 | 5 | 6 | 7 | 8 | 9 | 10 |
| 21, 22, 26 | 21, 22, 23 | 22, 23, 24 | 23, 24, 25 | 24, 25, 30 | 25, 26, 27 | 26, 27, 28 | 27, 28, 29 | 28, 29, 30 | 25, 29, 30 |

After training, the proposed method showed an accuracy of 0.966 on the test data set and 0.958 on the validation data set on average. Details of accuracy for each CNN on the mentioned data sets are presented in Table 2. The average accuracy in the test dataset is defined as the mean average of the network accuracy across all scenarios. It was observed that the average accuracy of the networks varies from 95.75% to 97.44% on the test data.

*Table 2; Accuracy of the CNNs on the training, test, and validation data sets.*

| | **Training** | **Test** | **Validation** |
|---|---|---|---|
| **Net ID** | Accuracy | Accuracy | Accuracy |
| **1** | 0.987 | 0.964 | 0.958 |
| **2** | 0.984 | 0.974 | 0.960 |
| **3** | 0.984 | 0.967 | 0.960 |
| **4** | 0.988 | 0.958 | 0.960 |
| **5** | 0.982 | 0.963 | 0.956 |
| **6** | 0.978 | 0.971 | 0.959 |
| **7** | 0.981 | 0.968 | 0.960 |
| **8** | 0.979 | 0.969 | 0.957 |
| **9** | 0.981 | 0.964 | 0.959 |
| **10** | 0.981 | 0.965 | 0.953 |

The DP values for each joint generated by the related CNN on the test data set are shown in Figures 3-5. In all 13 cases, damaged joints had a DP value greater than 96.6 percent, whereas undamaged joints had a DP value of less than ten.

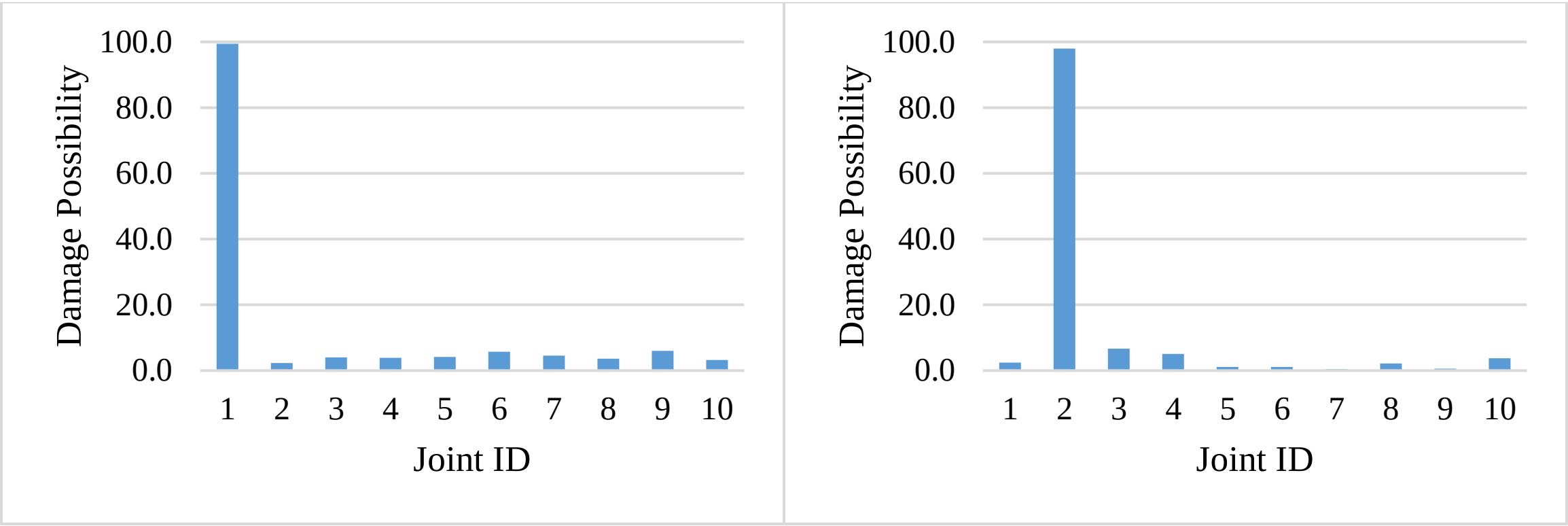

a b

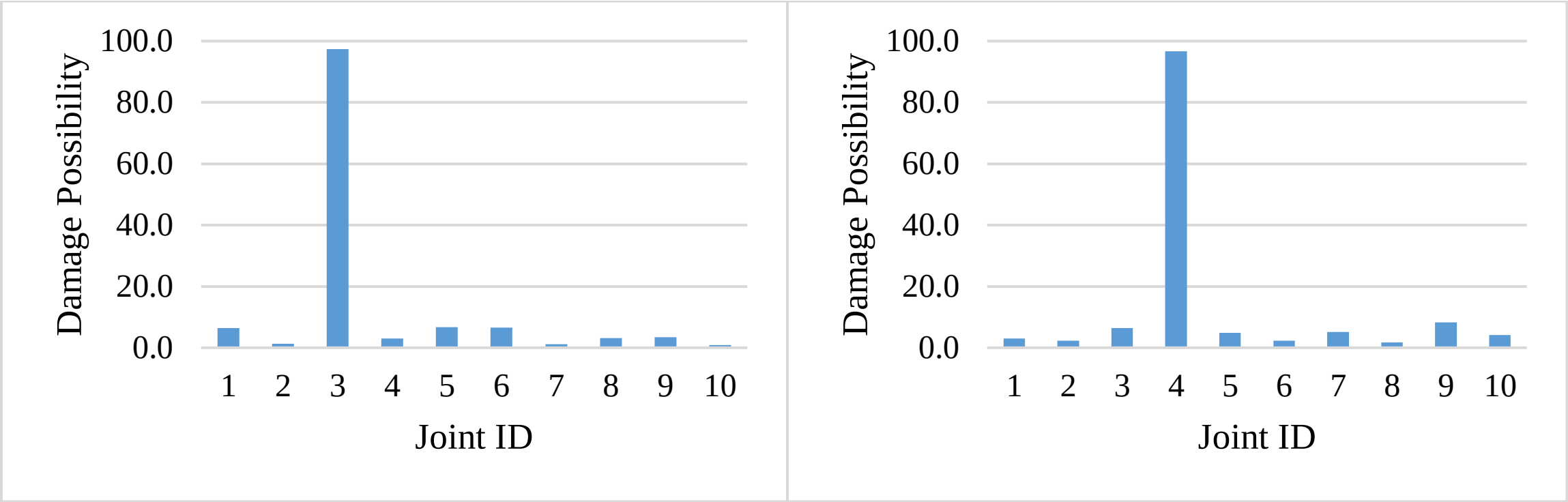

c d

*Figure 3; DP values for damage at a: joint 1, b: joint 2,c: joint 3, and d: joint 4*

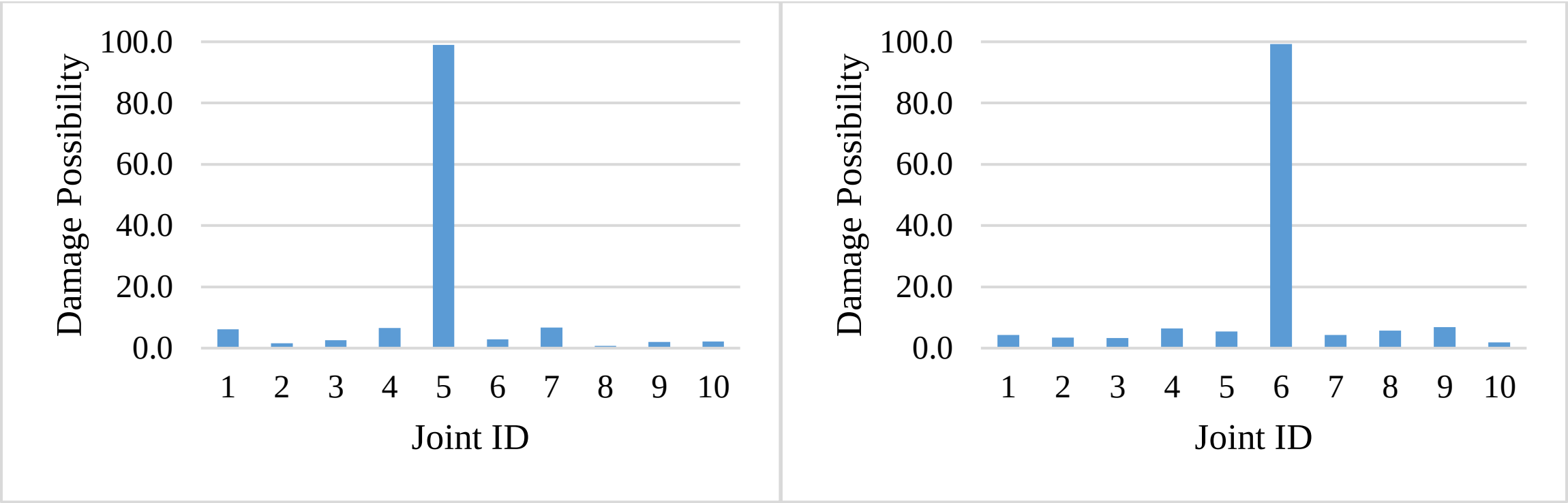

a b

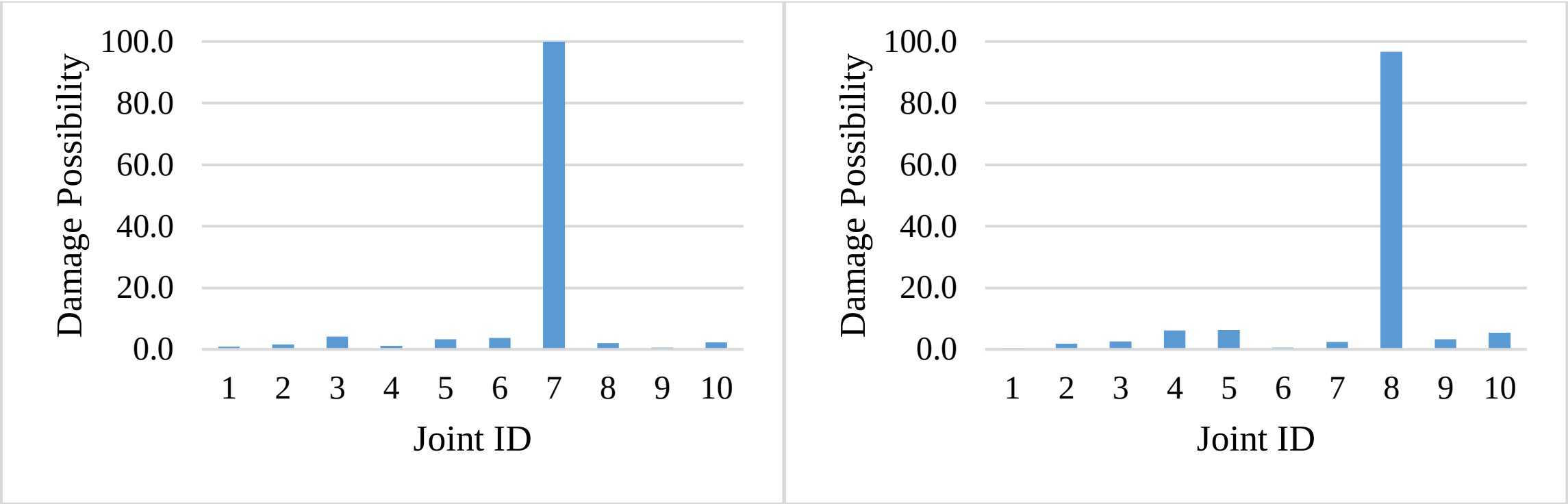

c d

*Figure 4; DP values for damage at a: joint 5, b: joint 6, c: joint 7, and d: joint 8*

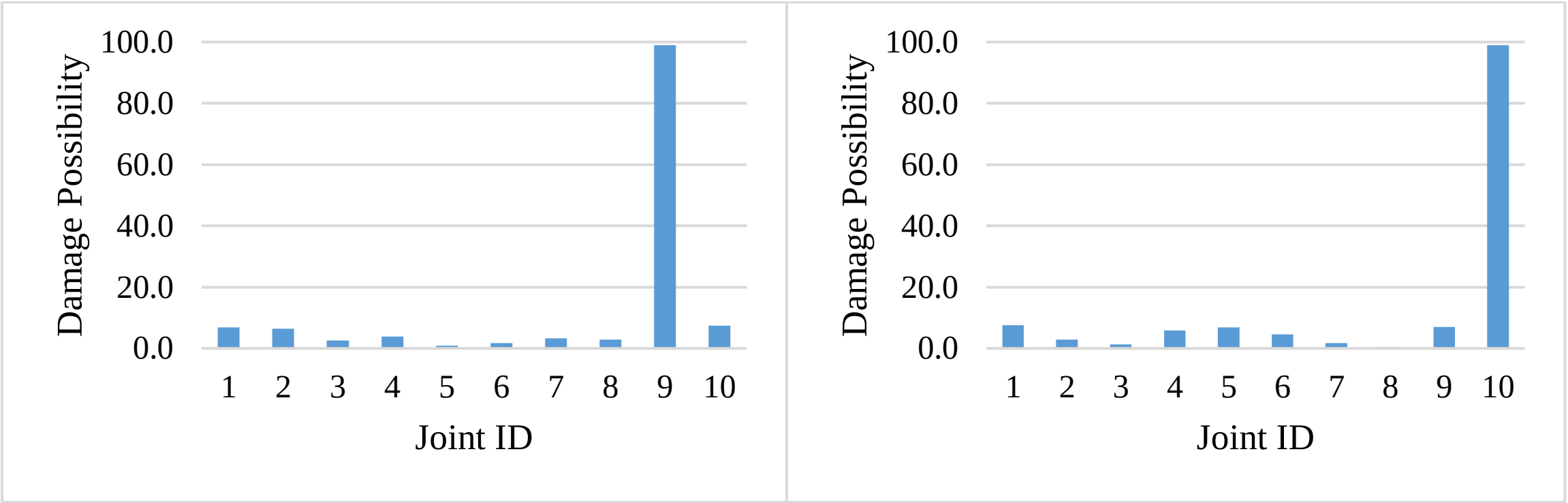

a b

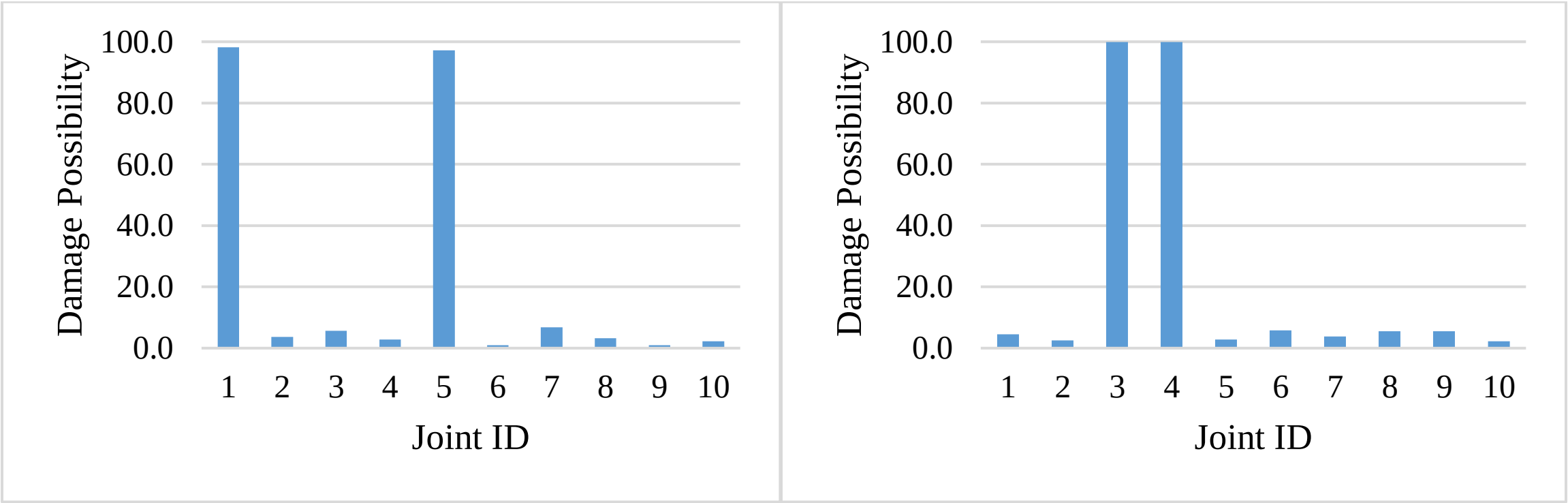

c d

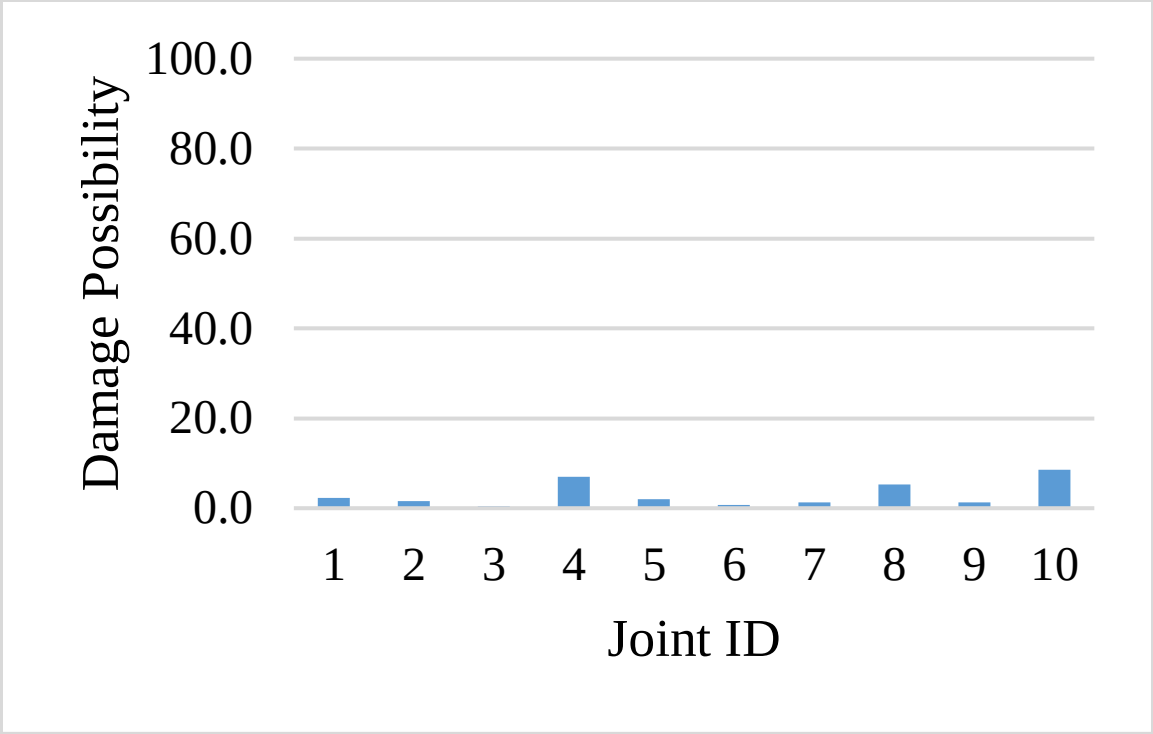

e

*Figure 5; DP values for a: damage at joints 9, b: damage at joint 10, c: damage at joints 1 and 5, d: damage at joints 3 and 4, and e: a fully undamaged case*

## 3.3 Speed Performance

A 32-second acceleration signal was sent to each network to determine the method's speed, and the required time was determined independently using the GPU and CPU. The results are an average of 50 runs. As indicated in Table 3, the average time required to process a 32-second signal is around 68 milliseconds on GPU and 73 milliseconds on CPU. Due to the compact architecture of the CNNs, the time required to process this signal on CPU and GPU was not considerably different. Thus, the time required to classify a one-second signal is around two milliseconds for each CNN, indicating that the method is fast and accurate, making it suitable for real-time methods.

*Table 3; required time (msec) for each CNN to process acceleration signals on both GPU and CPU*

| | | Network No. | | | | | | | | | | |
|---|---|---|---|---|---|---|---|---|---|---|---|---|
| | | 1 | 2 | 3 | 4 | 5 | 6 | 7 | 8 | 9 | 10 | average |
| **GPU** | 128 frames | 69 | 73 | 69 | 69 | 69 | 69 | 69 | 69 | 73 | 49 | 67.8 |
| | single frame | 0.5 | 0.6 | 0.5 | 0.5 | 0.5 | 0.5 | 0.5 | 0.5 | 0.6 | 0.4 | 0.5 |
| **CPU** | 128 frames | 78 | 82 | 82 | 71 | 78 | 74 | 80 | 76 | 82 | 49 | 75.3 |
| | single frame | 0.6 | 0.6 | 0.6 | 0.6 | 0.6 | 0.6 | 0.6 | 0.6 | 0.6 | 0.4 | 0.6 |

# 4 Conclusion

One of the most challenging aspects of SDD is patterning the measurements to the structural defects. This article addresses this issue by offering a framework for detecting and localizing structural deterioration that is both rapid and accurate, based on 2-D CNN. The suggested method is a decentralized approach in which each element is assigned a unique CNN. The method's applicability was investigated using the QUGS benchmark data set, and the following findings are possible:

- The experimental results show the high performance of the proposed method in detecting structural damages in both single and double damage cases.
- The algorithm is able to learn damage features directly from raw acceleration signals, requiring no preprocessing or handcrafted feature extraction.
- By adding signals of two more existing accelerometers to input, required data for training the CNNs is reduced. It also makes the algorithm able to employ 2-D CNN, which is faster in converge and more learnable than the 1-D type.
- Unlike traditional methods, this technique is decentralized, and each CNN follows a simple architecture. Accordingly, the proposed algorithm is computationally inexpensive and can be performed on several ordinary devices.
- As illustrated in section 3, the required time to assess new data is also less than needed for real-time approaches.

In commence, we observed ten joints in the benchmark as mentioned above to evaluate the proposed technique. As a result, the writers will endeavour to monitor the structure utilizing future research methods thoroughly. Additionally, the influence of additive noise on the training data should be investigated to test the networks' learnability in a practical simulation of real-world situations.

# 5 Acknowledgement

We would like to thank Dr Osama Abdeljaber from Structural Dynamics Team for kindly providing the data set we used in this article containing double damage cases.